\documentclass{article} 
\usepackage{iclr2021_conference,times}

\usepackage{amsmath,amsfonts,bm}

\def\eqref#1{equation~\ref{#1}}

\def\1{\bm{1}}

\DeclareMathAlphabet{\mathsfit}{\encodingdefault}{\sfdefault}{m}{sl}
\SetMathAlphabet{\mathsfit}{bold}{\encodingdefault}{\sfdefault}{bx}{n}

\usepackage{graphicx}
\usepackage{booktabs}
\usepackage{hyperref}
\usepackage{url}

\title{A learning gap between neuroscience\\ and reinforcement learning}

\author{Samuel T.\ Wauthier, Pietro Mazzaglia, Ozan Çatal, Cedric De Boom, Tim Verbelen \&\\
\bf Bart Dhoedt\\
IDLab, Department of Information Technology\\
Ghent University -- imec\\
Technologiepark-Zwijnaarde 126, B-9052 Ghent, Belgium\\
\texttt{\{samuel.wauthier, pietro.mazzaglia, ozan.catal, cedric.deboom, }\\
\texttt{tim.verbelen, bart.dhoedt\}@ugent.be}
}

\iclrfinalcopy 
\begin{document}

\maketitle

\begin{abstract}
Historically, artificial intelligence has drawn much inspiration from neuroscience to fuel advances in the field.
However, current progress in reinforcement learning is largely focused on benchmark problems that fail to capture many of the aspects that are of interest in neuroscience today.
We illustrate this point by extending a T-maze task from neuroscience for use with reinforcement learning algorithms, and show that state-of-the-art algorithms are not capable of solving this problem.
Finally, we point out where insights from neuroscience could help explain some of the issues encountered.
\end{abstract}

\section{Introduction}



Neuroscience and artificial intelligence have a long-standing history of cross-fertilization between the two domains~\citep{Hassabis2017}.
Most notable are advances in deep reinforcement learning (RL), which is heavily inspired by reward prediction error signals observed in the brain, as well as mimicking neural circuits for computation~\citep{Mnih2015DQNature}.
Current RL algorithms are even surpassing human performance on various, typically game environments such as Atari games~\citep{agent57}, Dota 2~\citep{Dota} or Go~\citep{alphago}.
However, these latest achievements are mainly attributed to incremental updates to the RL algorithm and scaling up training to huge amounts of data.

Progress in artificial intelligence is often driven by benchmark problems.
A clear example is the ImageNet data set, which is driving progress in computer vision~\citep{AlexNet,Microsoft,Khan2020}.
Also in RL, standard benchmarks were introduced to measure performance of various algorithms, such as the Atari Learning Environment~\citep{Atari}, Open AI Gym~\citep{Gym}, the DeepMind Lab~\citep{DeepMindLab} and Control Suite~\citep{DeepMindControl}.
Such benchmark problems become increasingly harder to solve, with the main focus being increasing the complexity of observations (i.e.\ from control states to pixels), and increasing the difficulty of finding rewards (i.e.\ sparse reward signals).
However, we argue that many aspects that are of interest in (behavioral) neuroscience, such as dealing with ambiguity, stochasticity, and memory, are less pronounced.

We propose that, once again, we should draw more inspiration from neuroscience and build novel benchmark problems that are, on the one hand, simple enough to enable relatively fast iterations for algorithm development, but on the other hand, hard enough for current algorithms to tackle.
One notable example in this direction is the Animal AI testbed~\citep{animalai}, which focuses on spatial navigation and tasks involving inference of object locations through object persistence and spatial elimination.
We believe that grounding RL benchmark problems in well-studied problems in neuroscience will also foster novel algorithms inspired by neuroscience theories and, at the same time, provide empirical evidence for these theories.



\begin{figure}[t]
\centering
\includegraphics[width=0.35\columnwidth]{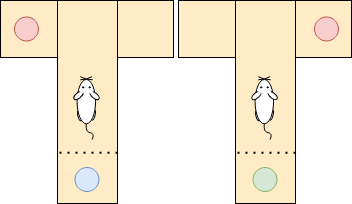}
\caption{Foraging environment as defined by \citet{Friston2016}. The left and right images represent two possible contexts. The agent starts in the middle. The red circle indicates the location of the reward. The green and blue circles are clues that tell the agent which context it is in, e.g.\ blue and green mean the reward will be on the left and right, respectively.}
\label{fig:friston}
\end{figure}

Psychology and neuroscience have been studying rodent behavior in mazes since the early 20th century~\citep{Tolman1930, OKeefe1971}.
A more recent example is the (artificial) T-maze, described by \citet{Friston2016}.
In this environment, an artificial agent, e.g.\ a rat, is put inside a T-shaped maze with a single reward, e.g.\ some cheese, in either the left or the right branch (see Figure \ref{fig:friston}).
The lower branch contains a clue on where the reward is located.
The agent starts in the center of the world and can choose which branch to take.
Crucially, after it has entered the left or right branch, it is not allowed to exit this branch.
The initial state of the world is ambiguous, since the agent has no way of knowing its context, i.e.\ which type of world it finds itself in: a world with a reward in the left branch or a world with a reward in the right branch.
In addition, without knowledge of whether the right or the left branch contains the reward, the agent can only guess which branch to take.
Observing the clue in the lower branch allows the agent to resolve the ambiguity and make an informed decision.
This T-maze was originally described in a discrete way, i.e.\ the agent can be in four possible locations (center, bottom, left, right) and has a maximum of two moves.
Note that the agent is allowed to go directly to the left or right branch from the bottom branch.
In their work, \citet{Friston2016} highlight that the key to solving the problem is the use of belief-based methods, whereas belief-free methods, such as dynamic programming, fail.

In the remainder of this paper, we revisit the environment described by \citet{Friston2016} and reimplement it as a pixel-based, game-like environment similar to the Atari benchmark.
Furthermore, we show that a set of state-of-the-art reinforcement learning algorithms is unable to solve this environment.
We conclude by discussing some of the limitations of current RL approaches and point out where inspiration from neuroscience might help.

\section{E-maze: T-maze revisited}

\begin{figure}[b]
\centering
\includegraphics[width=0.18\columnwidth]{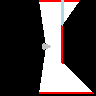}
\includegraphics[width=0.18\columnwidth]{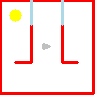}\hspace{1cm}
\includegraphics[width=0.18\columnwidth]{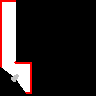}
\includegraphics[width=0.18\columnwidth]{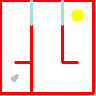}
\caption{(left) Starting position in E-maze (with and without black areas). (right) Path blocked after the agent enters a branch (with and without black areas).}
\label{fig:mouse}
\end{figure}

In this section, we extend the T-maze of \citet{Friston2016} in order to be compatible with current RL benchmarks\footnote{Environment source code is available at \url{https://github.com/thesmartrobot/ambigym}.}.
For example, we use pixel-based observations and allow the agent to move to intermediate positions between endpoints.
In other words, observations become images, and the state space becomes larger.
As a direct result, the maximum number of moves grows as well, since the environment cannot be solved in two moves anymore.

Our environment is shown in Figure~\ref{fig:mouse}.
It consists of a top-down view of a tilted E-shaped maze.
The grey triangle marks the agent, while the yellow circle indicates the reward.
Walls are displayed in red, and windows are displayed in blue.
Areas that the agent is unable to see are filled with black.
The agent can see through windows, but not through walls.
Figure~\ref{fig:mouse} also shows the environment with all the black-filled areas removed for the purpose of demonstration.
The agent's field of view (FOV) and maximum number of moves are hyperparameters.
Possible actions are: do nothing, move forward, and turn left or right by 45 degrees.
The moment the agent enters the left or right branch, the branch closes, and the agent cannot turn back (see Figure~\ref{fig:mouse}).

Note that the environment can be modeled as a partially observable Markov decision process (POMDP).
The agent is unable to infer the underlying state directly from a single observation, since areas that are not within the line of sight of the agent are invisible to the agent.
Moreover, from an RL standpoint, this setup corresponds to a sparse reward environment.
Indeed, no reward is gained unless the agent obtains the yellow circle.
Only if the agent reaches the yellow circle, it receives a reward of 1 and the episode is stopped.

\section{Experiments}
A number of experiments were set up to verify the performance of different RL algorithms on the environment.
The environment was initialized in the same manner for all experiments.
It was evaluated on model-free approaches (DQN~\citep{Mnih2015DQNature}, Rainbow~\citep{Rainbow}, PPO~\citep{PPO}), PPO with exploration bonuses (ICM~\citep{ICM}, RND~\citep{RND}) and a model-based approach (DreamerV2~\citep{DreamerV2}).
Details can be found in Appendix~\ref{app:algorithms} and~\ref{app:experiment}.

Performance was measured by how often the agent was able to reach the reward.
In practice, a rolling average of the returns over 100 episodes was used for the evaluation, where return denotes the episodic sum of rewards.
Since the return for this environment is either 0 or 1, the rolling average estimates the frequency with which the agent is able to reach the reward.
Furthermore, we average over 10 different training runs for each algorithm.
In what follows, we shall refer to `performance' to indicate the average number of times the agent was able to reach the reward.

\begin{figure}[t]
\centering
\includegraphics[width=0.9\textwidth]{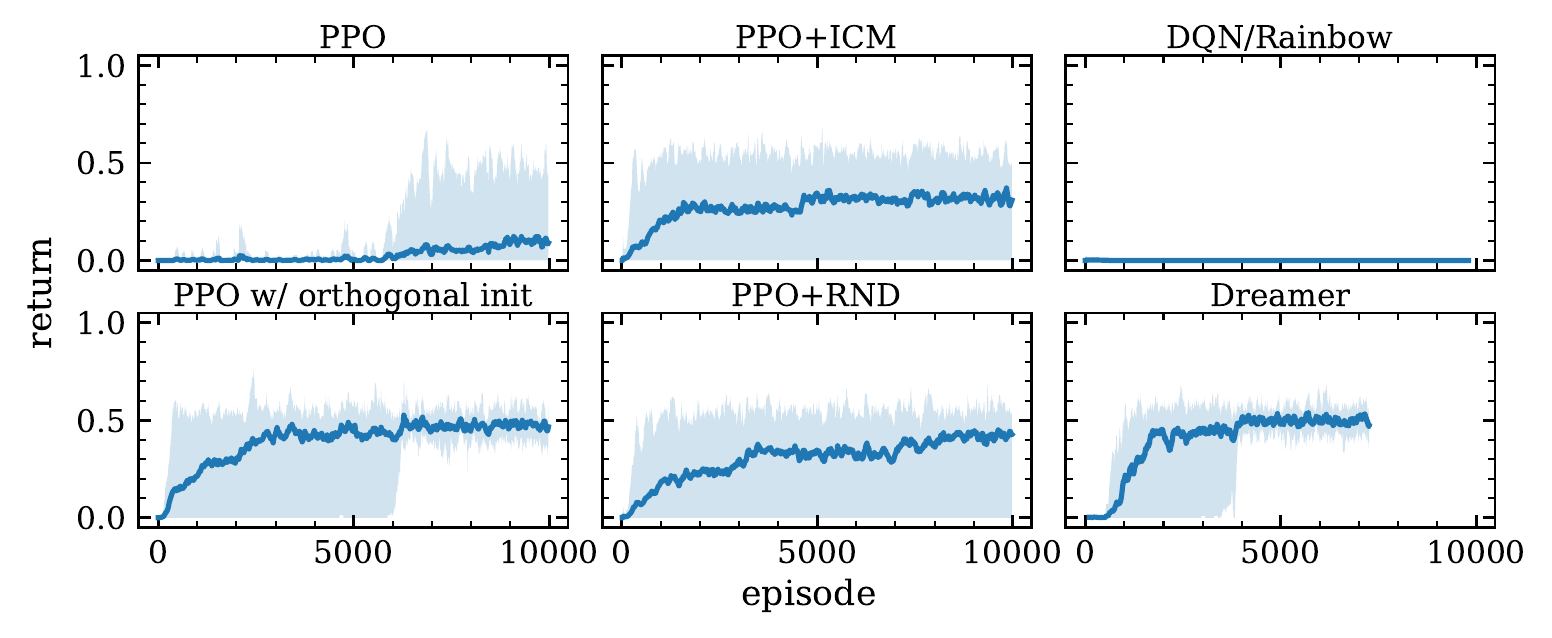}
\caption{Mean rolling average of episode return for different algorithms. Multiple runs were done for each algorithm and a rolling average was taken over 100 episodes for each run, after which the mean over all runs was taken. Shaded areas show the range between the minimum and maximum rolling average.}
\label{fig:results}
\end{figure}

Given the random nature of the environment, a random agent with an unbounded number of allowed moves would reach the reward 50\% of the time.
In other words, with the performance measure defined above, the algorithm under consideration must reach the reward more than 50\% of the time to be able to say that it performs better than a random agent.
This is important, since an RL agent is not guaranteed 50\% performance when the maximum number of moves is small.
For example, it may never learn to go towards the reward due to never reaching the reward.


Figure~\ref{fig:results} shows the mean rolling average of return per episode for the algorithms that learned to go to the reward over the course of training.
Shaded areas indicate the range between the minimum and maximum rolling average over all runs.
DQN and Rainbow displayed the same behaviour and are shown in the same plot.

Typical behavior of the rolling average curves of individual runs is that either the agent fails to learn and stays at 0\%, or the agent jumps to the level of a random agent from 0\% to around 50\%.
This is reflected in the minima and maxima.
For PPO and the exploration methods, whether the agent fails to learn depends on the seed.
As such, the mean over all runs reflects the amount of seeds for which the agent failed to learn.
For example, PPO reached a mean rolling average of 10\% over 10 runs, indicating that it failed 8 out of 10 times.

Out of the three model-free methods that were assessed, DQN and Rainbow fail to learn, while PPO (without orthogonal initialization or exploration) achieves very low performance.
The model-free methods suffer most from the sparse reward task and are unable to learn when the reward is not reached through random actions.
Adding exploration bonuses improves performance for PPO by 20--30\% on average.
Additionally, we found that using orthogonal weight initialization \citep{Saxe2013} drastically improves performance on PPO.
This is due to the fact that this yields more diverse initial random behavior.
Finally, DreamerV2 achieved 50\% after about 2000 episodes, but never increased to more than 50\%.

\section{Discussion}



The results show that model-free methods fail without specific initialization or exploration method.
On the one hand, this is a demonstration of how sensitive RL methods are to hyperparameters and how clever ``tricks'' must be used to obtain good results.
State-of-the-art results are usually obtained through important details, such as proper initialization, input normalization or adaptive learning techniques \citep{MakeRLWork}.
These details are often not highlighted as being crucial for the performance of the algorithm.
As a result, one could argue whether there is a default setting for these algorithms that generalizes well to any environment.
The right tuning is usually necessary to obtain good results.

On the other hand, methods that use a model, either for providing an exploration bonus or model-based approaches, at least succeed at reaching the reward and getting 50\% performance, regardless of initialization.
From a neuroscience perspective, one could point to theories in which the brain contains a model of the world, i.e.\ the brain maintains beliefs of the world through a generative model, such as Bayesian approaches to brain function \citep{BayesianBrain}, active inference \citep{Friston2006}, and the general adversarial brain \citep{Gershman2019}.
These models allow us to react to events with a stochastic nature and where information is not perfect.
Perhaps the use of a world model is crucial for algorithms to generalize well to any environment.





One important aspect of the environment is that rewards are sparse.
When rewards are sparse and difficult to find through random exploration, model-free approaches typically fail.
In that case, some incentive is needed to explore.
This incentive can be given through reward-shaping, intrinsic reward from exploration methods or, as mentioned earlier, clever initialization.
\citet{Motivation} argues that exploratory behavior in itself can be a source of reward.
For example, the theory of flow \citep{Flow} states that an important source of intrinsic reward for humans is the interest in activities which require slightly more skill than they currently possess.
Active inference \citep{Friston2006} frames the minimization of (variational) free energy as the intrinsic reward that governs the actions of living things.
The inclusion of an intrinsic reward, in addition to an extrinsic reward, may be a necessity.



Another important aspect of the environment is the relatively long time frame between seeing the reward through the window and obtaining the reward.
To optimally solve the E-maze, the agent will first need to go towards the windows to find out where the reward is, and then, remember the location when it chooses a branch.
Clearly, the agent will need some kind of memory to achieve this.
In some sense, model-based methods contain a type of implicit memory encoded in the state, though they can also use explicit memory, such as long short-term memory (LSTM), gated recurrent units (GRUs), memory networks~\citep{Weston2014} or neural Turing machines~\citep{Graves2014}.
However, as shown in the previous section, the model-based method DreamerV2 did not perform better than an agent which randomly chooses a branch without looking for the clue.
This may suggest the need to look to types of memory grounded in neuroscience, such as Hopfield networks~\citep{Hopfield} or Kanerva machines~\citep{Kanerva}.

Reinforcement learning needs to look to neuroscience for inspiration once again.
Current benchmarks are too restrictive in terms of the types of problems they address.
We introduced the E-maze, which was unable to be solved by state-of-the-art deep RL algorithms.
Finally, we argued how insights from neuroscience may aid in solving the E-maze.
The use of world models, intrinsic rewards, and memory, rooted in neuroscience, will be beneficial in this regard.


\subsubsection*{Acknowledgments}
This research received funding from the Flemish Government under the ``Onderzoeksprogramma Artifici\"ele Intelligentie (AI) Vlaanderen'' programme. Ozan Çatal is funded by a Ph.D grant of the Flanders Research Foundation (FWO).

\bibliography{iclr2021_conference}
\bibliographystyle{iclr2021_conference}

\newpage

\appendix
\section{Algorithm descriptions}\label{app:algorithms}

The following state-of-the-art RL algorithms were used to evaluate performance in our environment.

\textbf{DQN}\quad Deep Q Network (DQN; \cite{Mnih2015DQNature}) is a value-based optimization method.
The data collected from the environment is used to train a neural network predicting the action-value function $Q$, which provides estimates of the expected returns for all possible actions.
The policy is implemented as a selection process of the action with the highest $Q$ value at each time step.
For DQN, some exploration is provided through the $\epsilon$-greedy strategy, which consists of choosing a random action rather than the best one with probability $1 - \epsilon$.


\textbf{Rainbow}\quad Several improvements over the original DQN implementation have been combined in \citep{Rainbow}, leading to an overarching system referred to as `Rainbow'.
The three most important components of Rainbow are: the use of n-step returns \citep{Mnih2016NStepDQN}, the use of a prioritized experience replay \citep{Schaul2016PriorExpReplay} and modeling the Q function as a distribution \citep{Bellemare2017DistributionalRL}.

Both our DQN and Rainbow implementations rely on the Dopamine RL framework \citep{Castro2018Dopamine}, which provides state-of-the-art results on the Atari 2600 benchmark.

\textbf{PPO}\quad
The Proximal Policy Optimization (PPO; \cite{PPO}) algorithm is a policy-gradient method for deep RL.
Policy-gradient methods exploit the Policy Gradient theorem \citep{Sutton2018RL} to directly update the policy based on the environment returns.
With respect to Vanilla Policy Gradient (VPG or REINFORCE), PPO implements two additional aspects: a value function estimation, which is subtracted from the returns of the environment to reduce variance, and a clipped objective function, which prevents undesirable drastic updates of the policy.    

Our implementation of the PPO algorithm was based upon the original OpenAI Baselines implementation \citep{OpenAIBaselines}.


\textbf{Dreamer}\quad Dreamer \citep{Hafner2020Dreamer} is a model-based RL method based on two main principles: (i) learning a `world' model of the environment, which allows predicting future observations and rewards, (ii) using the world model to learn an optimal policy in an RL fashion, by applying policy-gradient updates over imaginary trajectories generated by the Dreamer's world model.

In our experiments, we ground on the second iteration of the Dreamer agent \citep{DreamerV2}, which was tested to surpass \citep{Rainbow} and other state-of-the-art model-free methods on the Atari 2600 benchmark.

\textbf{Exploration methods} Recent advances in exploration strategies for RL have shown significant performance improvements in sparse-reward tasks.
In the following, we present the two methods that we tested in our experimental setup:


\begin{itemize}
    \item \textit{ICM} - Intrinsic Curiosity Module \citep{ICM} consists of two networks that enable computing a curiosity bonus for exploration. The first is an inverse-dynamics model, used to learn compact features for reducing the dimensionality of the observation inputs. The second network is a forward-feature dynamics, which predicts the features of future observations by using the features of the current observation and the current action. The curiosity bonus is implemented as the error prediction of the forward-feature model, computed as the Mean Squared Error (MSE) between predictions and true features, and encourages the agent to explore different environment transitions.
    \item \textit{RND} - In Random Network Distillation \citep{RND}, they base upon previous work showing that random features can perform well to reduce the dimensionality of observations \citep{Burda2019LSCuriosity}. As a curiosity bonus, they use the MSE prediction error between features computed by a randomly-initialized network and a feature encoder they train, encouraging diversity in the environment observations.
\end{itemize}

In our experiments, we add the intrinsic bonuses computed by ICM and RND to the rewards, and train a PPO agent to maximize the compound returns.


\section{Experiment details}\label{app:experiment}


Since the environment is game-like, the decision was made to implement all algorithms with their original parameters used for the Atari benchmark.

Algorithms that are benchmarked with Atari 2600 usually use preprocessed images as input.
More specifically, observations from the Atari games are rescaled to 84x84 pixels and transformed to grayscale.
For this reason, images from our environment were preprocessed in the same manner.

The environment was initialized in the following way.
The agent was given an FOV of $1.1\pi$ radians (198 degrees) and a maximum amount of moves of 250.
The latter was chosen in a way to control the sparsity.
Too many moves could make the environment too easy to solve, so any random agent could obtain reward 50\% of the time.
Too few moves could make the environment too difficult to solve.
Since the minimum amount of moves necessary to solve the environment is around 50, the number was set to 250.

Important hyperparameters that may change between implementations in the literature are shown in Table~\ref{tab:runs}.
Parameters that were adjusted for our environment are emphasized.
Exploration methods were tested in combination with PPO.
The exploration algorithms were switched off after the agent reached 20\% average performance.
Aside from the default PPO initialization method, we ran a number of PPO experiments by initializing the policy network using large orthogonal weights.









\begin{table}[h]
\caption{Important hyperparameters in algorithm implementations. Some hyperparameters may change between implementation found in literature. This table indicates the exact parameters used. \emph{Emphasized text} indicates variables that were adjusted by us.}
\label{tab:runs}
\begin{center}
\begin{tabular}{lc}
\toprule
\multicolumn{2}{c}{\bf DQN}\\
\midrule
optimizer               & RMSProp\\
learning rate           & 2.5e-4\\
\emph{steps per batch}         & 2500\\
min.\ replay history    & 20000\\
batch size              & 32\\
\\
\toprule
\multicolumn{2}{c}{\bf Rainbow}\\
\midrule
optimizer               & Adam\\
learning rate           & 6.25e-5\\
\emph{steps per batch}         & 2500\\
min.\ replay history    & 20000\\
batch size              & 32\\
\\
\toprule
\multicolumn{2}{c}{\bf PPO}\\
\midrule
optimizer               & Adam\\
learning rate           & 2.5e-4\\
gradient clip           & 0.1\\
value loss coefficient  & 0.5\\
no. parallel agents     & 8\\
\emph{steps per batch}         & 250 (x8)\\
\emph{no. mini batches}        & 8\\
entropy coefficient     & 0.001\\
\\
\toprule
\multicolumn{2}{c}{\bf Exploration methods -- PPO}\\
\midrule
exploration off threshold               & 0.2\\
\bottomrule
\end{tabular}
\end{center}
\end{table}

\end{document}